\documentclass{fcs}
\usepackage{bm}
\usepackage{algorithm} 
\usepackage{algorithmic} 
\usepackage{caption}
\usepackage{times}
\usepackage{epsfig}
\usepackage{graphicx}
\usepackage{amsmath}
\usepackage{amssymb}

\definecolor{Revision}{rgb}{0,0,0}

\volumn{ }
\doi{ }
\articletype{RESEARCH~ARTICLE}
\copynote{{\copyright} Higher Education Press and Springer-Verlag Berlin Heidelberg 2016}
\ratime{Received month 02, 2016; accepted month 08, 2016}
\email{$sgshan@ict.ac.cn$}
\title{VIPLFaceNet: An Open Source Deep Face Recognition SDK}
\author{Xin Liu$^{1,2}$, Meina Kan$^{1,2}$, Wanglong Wu$^{1,2}$, Shiguang Shan \xff $^{1,2}$, Xilin Chen$^{1,2}$}
\address{{1\quad Key Lab of Intelligent Information Processing of Chinese Academy of Sciences (CAS),\\
 Institute of Computing Technology, CAS, Beijing, 100190, China}\\
{2\quad University of Chinese Academy of Sciences, Beijing 100049, China}}

\begin{document}
\maketitle
\setcounter{page}{1}
\setlength{\baselineskip}{14pt}

\begin{abstract}
Robust face representation is imperative to highly accurate face recognition. In this work, we propose an open source face recognition method with deep representation named as VIPLFaceNet, which is a 10-layer deep convolutional neural network with 7 convolutional layers and 3 fully-connected layers. Compared with the well-known AlexNet, our VIPLFaceNet takes only 20\% training time and 60\% testing time, but achieves 40\% drop in error rate on the real-world face recognition benchmark LFW. Our VIPLFaceNet achieves 98.60\% mean accuracy on LFW using one single network. An open-source C++ SDK based on VIPLFaceNet is released under BSD license. The SDK takes about 150ms to process one face image in a single thread on an i7 desktop CPU. VIPLFaceNet provides a state-of-the-art start point for both academic and industrial face recognition applications.

\end{abstract}


\section{Introduction}
\noindent Face recognition, as one of the typical problems in computer vision and machine learning, plays an important role in many applications, such as video surveillance, access control, computer-human interface and mobile entertainments \cite{Survey}. Generally speaking, a conventional face recognition system consists of four modules, face detection, face alignment, face representation and identity classification. In this pipeline, the key component for accurate face recognition is the third module, i.e. extracting the representation of an input face, which this paper mainly focuses on.

The main challenges of face representation lie in the small inter-person appearance difference caused by similar facial configurations, as well as the large intra-person appearance variations due to large intrinsic variations and diverse extrinsic imaging factors, such as head pose, expression, aging, and illumination. In the past decades, face representation is mostly based on hand-crafted local descriptors \cite{GaborFace,LBPFace,HighdimLBP,HOGFace01,POEM,LQP,SIFTFace01} and shallow learning-based representation models \cite{kumar2012trainable,LeiZhen,LearnGaborFace,LE2010,SFRD,TomvsPete}.
As the development of deep learning technology, it becomes a more potent approach for face representation learning, especially in the real-word scenarios. Compared with the previous hand-crafted routine, deep face representation is learned in a data-driven style which can guarantee better performance as validated in \cite{Deepface,DeepID,DeepID2,DeepID2+,FaceNet}. Taking the de-facto real-world face recognition benchmark LFW as an example, hand-crafted descriptor recorded 95.17\% set by high-dimensional LBP \cite{HighdimLBP}, while 99.63\% accuracy achieved by the latest deep FaceNet in \cite{FaceNet}.

In spite of many decades of research and development on face recognition, few open-source face recognition systems are publicly available yet. An open-source SDK with high accuracy in general scenarios is in great need for both academic research and industrial applications. So in this work, we meet this requirement and propose a deep face recognition model named as VIPLFaceNet, which is released as a BSD-license open source software with detailed implementation of the recognition algorithm. VIPLFaceNet is a powerful deep network for face representation with ten layers including 7 convolutional layers and 3 fully-connected layers. As a BSD-license open source software, VIPLFaceNet allows both academic research and industrial face recognition applications in different software and hardware platforms for free.

The contributions of this paper are summarized as follows:

1. We propose and release an open source deep face recognition model, VIPLFaceNet, with high-accuracy and low computational cost, which is a 10-layer deep convolutional neural network that achieves 98.60\% mean accuracy on the real-world face recognition benchmark LFW.

2. We investigate the network architecture design and simplification. By careful design, VIPLFaceNet reduces 40\% computation cost and cuts down 40\% error rate on LFW compared with the AlexNet \cite{DCNN}.

3. The VIPLFaceNet SDK code is written in pure C++ code under the BSD license. It is free and easy to be deployed in various software or hardware platforms for both academic research and industrial face recognition applications.

In summary, VIPLFaceNet is an open source deep face recognition SDK with high accuracy in general scenarios, which is built for facilitating the academic and industrial application of various real-world face recognition tasks. The rest of this paper is organized as follows. Section 2 presents the related works on face representation learning and introduce the face recognition benchmarks. Section 3 presents the network architecture design and technical details of our VIPLFaceNet. Section 4 conducts the experimental evaluation with comprehensive discussions and section 5 concludes this paper.

\section{Related Works}
In this section, we give a brief review of the related works on face representation learning. Moreover, we give a brief review
of the face recognition benchmarks and discuss the performance evolution on the de-facto real-world face recognition benchmark LFW.

\subsection{Face Representation before Deep Learning}
In the past decades, numerous hand-crafted local features were proposed for face representation, e.g. Gabor wavelets \cite{GaborFace}, Local Binary Pattern (LBP) \cite{LBPFace} and its high dimensional variant \cite{HighdimLBP}, Scale-Invariant Feature Transform (SIFT) \cite{SIFTFace01}, Histogram of Oriented Gradients (HOG) \cite{HOGFace01}, patterns of oriented edge magnitudes (POEM) \cite{POEM}, Local Quantized Pattern (LQP) \cite{LQP} etc. However, designing an effective local descriptor demands considerable domain specific knowledge and a great deal of efforts.

Besides the hand-crafted local features, learning-based representation is also popular and reports promising accuracy. In \cite{kumar2012trainable} and \cite{LeiZhen}, filters are learned to maximize the discriminative power for face recognition. In \cite{OBF}, faces are represented from its responses to many pre-trained object filters. In \cite{LearnGaborFace}, \cite{LE2010}, \cite{SFRD} and \cite{FishervectorFace}, codebook learning technologies are utilized for robust face representation. More recently, faces are represented with mid-level or high-level semantic information. For instance, the attributes and simile classifier \cite{kumar2009attribute} represent faces by the mid-level face attributes and so-called simile feature. Tom-vs-Pete classifier \cite{TomvsPete} encodes faces with high-level semantic information by the output scores of a large number of person-pair classifiers. Different from the deep learning approaches, the above methods are still shallow models and mostly rely on hand-crafted local features.

\subsection{Deep Face Representation Learning}
In recent years, deep learning methods are exploited to learn hierarchical representation and report state-of-the-art performance on LFW \cite{Deepface,DeepID,DeepID2,DeepID2+,CASIA-Web,FaceNet}.

DeepFace is an early attempt of applying deep convolutional neural network in real-world face recognition. There are four highlights in DeepFace: 1) A 3D model based face alignment to frontalize facial images with large pose. 2) A very large scale training set with 4 million face images of 4,000 identities. 3) Deep convolutional neural network with the local connected layer that learns separate kernel for each spatial position. 4) A Siamese network architecture to learn deep metric based on the features of the deep convolutional network.

The DeepID \cite{DeepID}, DeepID2 \cite{DeepID2} and DeepID2+ \cite{DeepID2+} are a series of works, which provide a very good example of deep network evolution. In DeepID, 25 CNN networks are trained on each face patch independently. Besides, Joint Bayesian method\cite{JointBayesian} is applied to learn robust face similarity metric. Finally, an ensemble of 25 deep networks achieve 97.45\% mean accuracy on LFW. The DeepID2 introduces the joint identification and verification losses. The performance of DeepID2 on LFW is improved to 99.15\%. The DeepID2+ just makes the network deeper and adds auxiliary loss signal on lower layer. Besides, the activation of the feature embedding layer is also studied as sparse, selective and robust. The mean accuracy of DeepID2+ on LFW is 99.47\% with 25 CNN models.

Learning face representation from scratch \cite{CASIA-Web} presents a semi-automatic way to collect face images from the internet and builds a large scale dataset CASIA-Web containing about 494,414 images of 10,575 subjects. Then a 13-layer deep network with 10 convolutional layers and 3 fully-connected layer is trained with joint identification and verification losses, reporting 97.73\% accuracy on LFW.

Another promising deep neural network is FaceNet \cite{FaceNet} proposed by Google, which uses a super large scale face dataset containing 200 million face images of 8 million face identities to train a GoogLeNet network. Given such a large number of identities, the classical Softmax loss which needs the same number of 8 millon output nodes consumes too much GPU memory. So instead, a triplet loss which does not consume extra memory is introduced in FaceNet to directly optimize the embedding feature and achieves 99.63\% mean accuracy on LFW.

All the above deep learning methods achieve quite promising face recognition accuracy on the challenging LFW dataset. The superiority demonstrates the superiority of the favorable feature learning ability of the deep neural networks.

\subsection{Evolution of Benchmarks}

In early years, most datasets for face recognition were collected in controlled environment, e.g. ORL \cite{ORL}, AR \cite{AR}, FERET \cite{FERET}, PIE \cite{PIE}, FRGC \cite{FRGC}, Extended-Yale-B \cite{EYaleB}, CAS-PEAL \cite{CASPEAL} and MultiPIE \cite{MultiPIE}. Among these datasets, AR is specially regarded as a benchmark to study occlusion robust face recognition. FERET, CAS-PEAL, PIE and MultiPIE are often used as general benchmarks to evaluate different factors of face recognition, such as aging, pose, expression, illumination, accessories etc. Yale-B is often cited as a lighting robust face recognition benchmark. Among these datasets, FRGC is widely used as a more challenging face recognition benchmark as it consist of over 50,000 images collected in varying lighting condition, e.g. atria, hallways, or outdoors.

In recent years, lots of real-world face datasets have been released, such as Labeled Face in the Wild \cite{LFW}, PubFig \cite{kumar2009attribute}, CelebFaces \cite{DeepID}, WDRef \cite{JointBayesian}, SFC \cite{Deepface}, CACD \cite{CACD}, WLFDB \cite{wlfdb2014}, CASIA-Web \cite{CASIA-Web}, MSRA-CFW \cite{MSRA-CFW} etc. These datasets are built for different motivations. Among them, CelebFaces, WDRef, SFC and CASIA-Web are built to train model for testing on LFW \cite{LFW}, but only CASIA-Web is a public dataset. WLFDB is a weakly labeled dataset without accurate identity annotation and acts as a search-based face tagging benchmark. CACD is built for studying cross-age face recognition, but only 10\% of the subjects in CACD are manually annotated. The MSRA-CFW dataset is built for face retrieval evaluation and the face identity is automatically annotated by algorithms.

\begin{figure}[t]
\centering
\includegraphics[width=8cm]{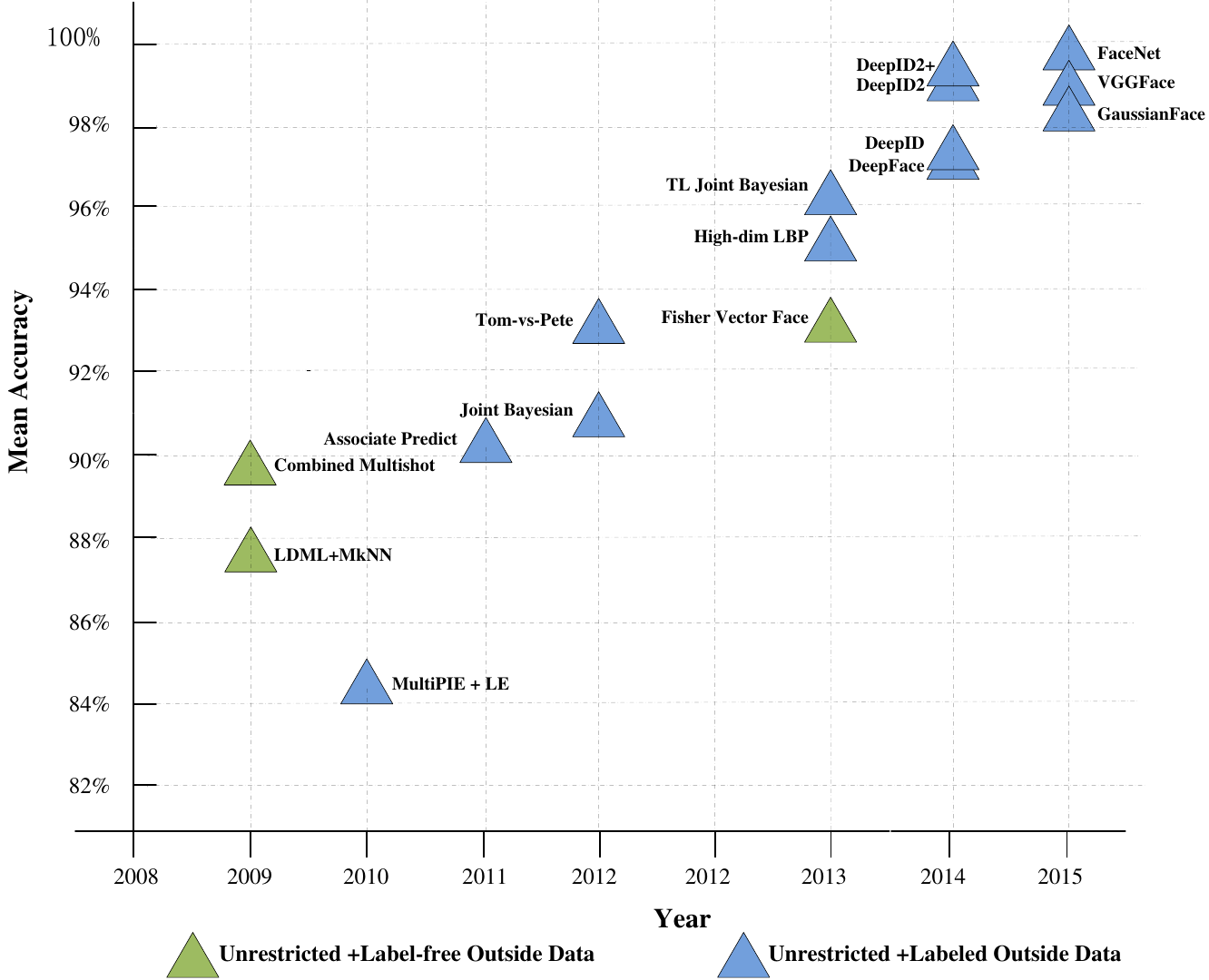}
\caption{Evolution of the face recognition techniques on LFW from 2009 to 2015. In the figure, triangles with different color refer to what type of outside training data is used in the experiments.}
\label{fig:lfwevo}
\vspace{-15pt}
\end{figure}

In the past several years, LFW has become the de-facto benchmark for real-world face recognition. According to the standard LFW protocol, the performance measurement should be the mean accuracy over 10-fold face verification task with each fold containing 300 inter-class and 300 intra-class face pairs. Besides the standard verification protocol, a face identification protocol is also available in \cite{LFWOpenSet}. A brief history of the performance evolution of LFW is demonstrated in Figure \ref{fig:lfwevo}. Representative methods including LDML-MkNN \cite{LDML}, Multishot\cite{multipleshot}, LE \cite{LE2010}, Associate-Predict \cite{AP2011}, Tom-vs-Pete\cite{TomvsPete}, Fisher Vector Face\cite{FishervectorFace}, High-dim LBP\cite{HighdimLBP}, TL Joint Bayesian\cite{TLJB2013}, DeepFace\cite{Deepface}, DeepID\cite{DeepID}, DeepID2\cite{DeepID2},
Gaussian Face\cite{GaussianFace}, VGGFace\cite{VGGFace}, DeepID2+\cite{DeepID2+} and FaceNet\cite{FaceNet} are shown.

\begin{table*}
\caption{Comparisons of AlexNet \cite{DCNN}, VIPLFaceNetFull and VIPLFaceNet. In the table, S denotes stride, G denotes group and Pad denotes padding. The ReLU layers are not shown in this table for the efficiency of presentation.}
\vspace{-10pt}
\label{table:netarc}
\begin{center}
\begin{tabular}{lll} \hline
AlexNet                          &VIPLFaceNetFull                  & VIPLFaceNet                    \\ \hline
Conv1: 96x11x11, S:4, Pad:0      &Conv1: 96x9x9, S:4, Pad:0        &Conv1: 48x9x9, S:4, Pad:0           \\
LRN                              &--                               &--         \\
Pool1: 3x3,S:2                   &Pool1: 3x3,S:2                   &Pool1: 3x3,S:2                      \\
Conv2: 256x5x5, G:2, S:1, Pad:2  &Conv2: 192x3x3,S:1, Pad:1        &Conv2: 128x3x3,S:1, Pad:1           \\
LRN                              &--                               &--         \\
--                               &Conv3: 192x3x3,S:1, Pad:1        &Conv3: 128x3x3,S:1, Pad:1           \\
Pool2: 3x3,S:2                   &Pool2: 3x3,S:2                   &Pool2: 3x3,S:2                      \\
Conv3: 384x3x3, S:1, Pad:1       &Conv4: 384x3x3,S:1, Pad:1        &Conv4: 256x3x3,S:1, Pad:1           \\
Conv4: 384x3x3, G:2, S:1, Pad:1  &Conv5: 256x3x3,S:1, Pad:1        &Conv5: 192x3x3,S:1, Pad:1           \\
--                               &Conv6: 256x3x3,S:1, Pad:1        &Conv6: 192x3x3,S:1, Pad:1           \\
Conv5: 256x3x3, G:2, S:1, Pad:1  &Conv7: 192x3x3,S:1, Pad:1        &Conv7: 128x3x3,S:1, Pad:1           \\
Pool3: 3x3,S:2                   &Pool3: 3x3,S:2                   &Pool3: 3x3,S:2                      \\
FC1, 4,096                       &FC1, 4,096                       &FC1, 4,096                           \\
Dropout1: dropout\_ratio:0.5     &Dropout1: dropout\_ratio:0.5     &Dropout1: dropout\_ratio:0.5        \\
FC2, 4,096                       &FC2, 2,048                       &FC2, 2,048                           \\
Dropout2: dropout\_ratio:0.5     &Dropout2: dropout\_ratio:0.5     &Dropout2: dropout\_ratio:0.5        \\
FC3, 10,575                      &FC3, 10,575                      &FC3, 10,575                         \\\hline
\end{tabular}
\end{center}
\vspace{-15pt}
\end{table*}

Figure \ref{fig:lfwevo} illustrates the evolution of face recognition techniques along with the accuracy increases: from early hand-crafted local features to shallow representation learning until recent deep representation learning. The first performance breakthrough is made by LDML-MkNN \cite{LDML}, which is a metric learning method, then the representation learning becomes the engine of accuracy improvement \cite{multipleshot,TomvsPete,HighdimLBP,FishervectorFace}. Finally, deep learning approaches reach the best results on LFW \cite{Deepface,DeepID,DeepID2,DeepID2+,VGGFace,FaceNet}.

\section{Proposed VIPLFaceNet}
This section presents the details of our proposed VIPLFaceNet. Firstly, we introduce the network architecture design and simplification. Then, to accelerate the training of the deep network, we introduce the fast normalization layer. Finally, we present the technical details of face pre-processing and deep network training.

\subsection{Network Architecture}
The network architecture is the essential part of a deep model. Recently, some network architectures has been well recognized, such as AlexNet \cite{DCNN}, GoogLeNet \cite{GoogLeNet} and VGGNet \cite{VGG}. Among them, the AlexNet is the simplest, with 5 convolutional layer and 3 fully-connected layers \cite{DCNN}. Besides the convolutional layer and fully-connected layer, the ReLU layer and the dropout operation, proposed in AlexNet, build the basis of the latest deep convolutional neural networks. Another important component of AlexNet is its local response normalization layer (LRN) which can improve the generalization ability of AlexNet. Aiming to go deeper, GoogLeNet is designed to be a 22-layer deep network and introduces an inception structure which extracts multi-scale features \cite{GoogLeNet}. Differently, the VGGNet uses only $3\times3$ convolutional kernel and the stride is always set to 1 \cite{VGG}, while the feature map size is reduced only by the pooling operation. Among the above three networks, VGGNet is the slowest, while AlexNet is the simplest.

\textbf{Design and simplification of VIPLFaceNet Network}. Considering the success and efficiency of AlexNet, our network is designed by adapting AlexNet to incorporate some recent new findings. Compared with AlexNet, our VIPLFaceNet design has six main features: 1) we use $9\times9$ size for the first convolutional layer rather than $11\times11$, to reduce the computational cost. 2) We remove all local response normalization layers, as we found it unnecessary provided proper parameter initialization \cite{PReLU}. 3) we decompose the second $5\times5$ convolutional layer of AlexNet to two $3\times3$ layers, inspired by He \textit{et al.}'s work\cite{CNNCTC}. 4) Specially, we remove all group structures in AlexNet as we exploit a more efficient way to do parallel training, i.e. asynchronous stochastic gradient descent \cite{ASGD}. 5) Further, we reduce the number of feature maps in each layer and add one more convolutional layer. 6) The number of nodes in the FC2 fully-connected layer is reduced to 2,048 from 4,096 inspired by the experimental analysis in \cite{chatfield2014return}.

The above six features lead to our VIPLFaceNetFull network consisting of 7 convolutional layers followed by 3 fully-connected layers, as shown in Table \ref{table:netarc}. Its computational cost is almost 90\% of AlexNet, which is still very high. To further reduce the computational cost, we simplified VIPLFaceNetFull to VIPLFaceNet by reducing the number of filters in the convolutional layers, as detailed in Table \ref{table:netarc}. Eventually, the VIPLFaceNet network consumes only 60\% computations of AlexNet.

\subsection{Fast Normalization Layer}
As shown in LeCun \textit{et al.}'s work \cite{EBP}, data normalization can speed up convergence, which is recently extended as the batch normalization
algorithms \cite{BN}. Inspired by these works, we also exploit a fast normalization layer in our VIPLFaceNet before the ReLU layer to speed up the convergence and improve the generalization.

Specifically, the fast normalization layer aims at normalizing the output of each network node to be of zero mean and unit variance. {\color{Revision}Unlike the batch normalization in \cite{BN}, our fast normalization layer does not have the recovery operation and thus consumes less GPU memory and computation cost}. Suppose the output of the network consists of $C$ dimensions, and the normalization is applied to each dimension independently. Next we take the $k$-th dimension as an example for illustrating and omit $k$ for simplicity. The $k$-th dimension of the network output for all $N$ training samples in a mini-batch is denoted as $\mathcal{B}_x = {x_1,x_2,...,x_N}$. We denote the fast normalization layer (FNL) as:

\begin{equation}\label{NL}
  FNL: x_1,x_2,...,x_N \rightarrow \hat{o}_1,\hat{o}_2,...,\hat{o}_N, \forall i, \hat{o}_i\sim N(0,1),
\end{equation}

\noindent{where $N(0,1)$ denotes the standard normal distribution with zero mean and unit variance. We present the detail of fast normalization layer (FNL) in Algorithm \ref{alg:Norm}. In the algorithm, $\mu_x$ is initialized as $0$ and $\sigma_x$ is initialized as $1$, and $\omega$ is the momentum value and set as $0.99$ by default. In the test phase, $\mu_x$ and $\sigma_x$ obtained in the final training stage are directly adopted.}

During training, the fast normalization layer backpropagates the gradient using the chain rule as follows:

\begin{equation}\label{BP}
  \begin{aligned}
   \frac{\partial L}{\partial \sigma} &= -\frac{1}{2}\sum_{i=1}^{N}\frac{\partial L}{\partial \hat{o}_i}(x_i - \mu)\sigma^{-3/2}.\\
   \frac{\partial L}{\partial \mu} &= (\sum_{i=1}^{N}\frac{\partial L}{\partial \hat{o}_i}\frac{-1}{\sqrt{\sigma}})+\frac{\partial L}{\partial \sigma}
   \frac{-2\sum_{i=1}^{N}(x_i - \mu)}{N}.\\
   \frac{\partial L}{\partial x_i} &= \frac{\partial L}{\partial \hat{o}_i}\frac{1}{\sqrt{\sigma}} +
   \frac{\partial L}{\partial \sigma}\frac{2(x_i - \mu)}{N} + \frac{1}{N}\frac{\partial L}{\partial \mu}.  \\
   \end{aligned}
\end{equation}

\begin{algorithm}[htb] %
\renewcommand{\algorithmicrequire}{\textbf{Input:}}
\renewcommand\algorithmicensure {\textbf{Output:} }
\caption{Fast Normalization Layer (FNL)} %
\label{alg:Norm} %
\begin{algorithmic}[1] %

\REQUIRE DCNN Network and mini-batch $\mathcal{B}_x$\\
\ENSURE Normalized output for each sample in $\mathcal{B}_x$ \\
\STATE Calculate the batch mean: $\mu = \frac{1}{N} \sum\limits_{i = 1}^N x_i$
\STATE Calculate the batch variance: $\sigma = \frac{1}{N} \sum\limits_{i = 1}^N (x_i - \mu)^2$
\STATE Calculate the normalized value: $\hat{o}_i = \frac{x_i - \mu}{\sqrt{\sigma}}$
\STATE Update the global mean: $\mu_x = \omega * \mu_x  + (1-\omega)*\mu$
\STATE Update the global variance: $\sigma_x = \omega * \sigma_x  + (1-\omega)*\sigma$.

\RETURN $\hat{o}_i$, i = 1, 2,..., $N$.
\end{algorithmic}
\end{algorithm}

Additionally, as observed from extensive experiments, the dropout operation can be safely removed for deep network with fast normalization layer. It is observed that not only the deep network training is greatly accelerated but also the generalization ability is improved. In the experimental sections, we will validate the effectiveness of the fast normalization layer.

\subsection{Technical Details}
In all experiments, the face images are preprocessed with three steps including face detection, facial landmark localization and face normalization.

\textbf{Face Detection}: In face detection stage, we adopt the face detection toolkit developed by VIPL lab of CAS \cite{VIPL}. One can refer to \cite{VIPLFaceDetect} for more details.

\textbf{Facial Landmark Localization}: We apply the Coarse-to-Fine Auto-Encoder Networks (CFAN) \cite{CFANToolbox} to detect the five facial landmarks in the face, i.e. the left and right center of the eyes, the nose tip, the left and right corner of mouth.

\begin{figure}[t]
\centering
\includegraphics[width=8cm]{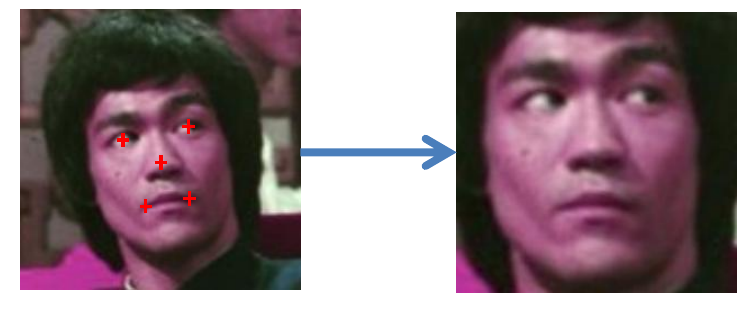}
\caption{Example of face normalization using five points.}
\label{fig:facetemplate}
\end{figure}

\textbf{Face Normalization}: As shown in Figure \ref{fig:facetemplate}, the face image is normalized to $256\times256$ pixels using five facial landmarks.

\textbf{Training details} In all experiments, for those deep networks without fast normalization layer , the base\_lr is set as 0.01 and the learning rate is reduced following polynomial curve with gamma value equal to 0.5. For those deep networks with the fast normalization layer, the base\_lr is set as 0.04. The momentum is set as $0.9$ and the weight decay is set as $0.0005$. All the experiments are conducted in Titan-X GPU with 12GB memory using a modified Caffe deep learning toolbox \cite{jia2014caffe}.

\begin{figure*}[t]
\centering
\includegraphics[width=12cm]{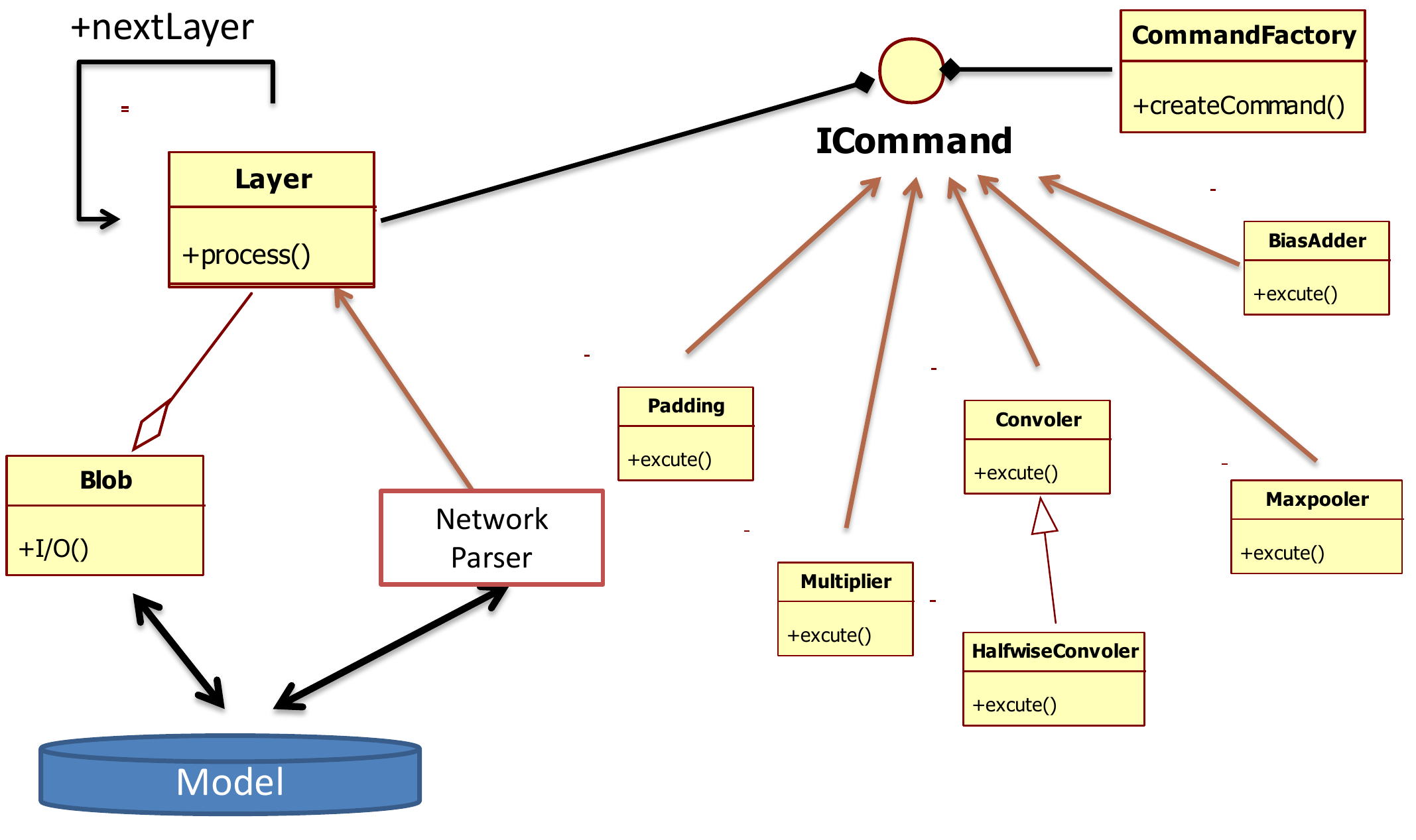}
\caption{The VIPLFaceNet SDK Architecture.}
\label{fig:sdkarc}
\end{figure*}

\section{SDK Architecture}
As mentioned previously, the source code of our VIPLFaceNet is released under the BSD license, and now available on https://github.com/seetaface. In this section, we introduce the highlights and architecture of the SDK for a better understanding of the source code.

\subsection{Highlights of VIPLFaceNet SDK}
This open-source SDK provides a powerful toolkit for testing and deploying face recognition applications, setting a good starting point for researchers and developer to experience the state-of-the-art face recognition technology.

\textbf{High-performance}. The VIPLFaceNet achieves state-of-the-art performance on the LFW benchmark with only single network. {\color{Revision}Further performance improvement can be expected by using metric leaning approaches, e.g. Joint Bayesian method \cite{JointBayesian} and MRMD \cite{RankingMetric} or classifier ensemble approaches, e.g. LibD3C \cite{LibD3C}.}

\textbf{Object-oriented}. The VIPLFaceNet SDK is designed from the beginning to be an object-oriented software, allowing easy extension to new network layers and any user-defined network architecture. It can also be easily integrated into industrial face recognition systems for various tasks.

\textbf{Configurable Network Architecture.} VIPLFaceNet SDK facilities the network architecture configuration independent of the SDK code. The network definitions is saved in the model file using pre-defined format. VIPLFaceNet supports network architectures in the form of arbitrary directed acyclic graphs. Upon initialization, VIPLFaceNet parses the network architecture from the model file and loads the network parameters into memory.

\textbf{Pure C++ Code}. The VIPLFaceNet is implemented in fully C++ code. It is very efficient to deploy VIPLFaceNet in multiple hardware platforms and operation systems.

\textbf{Community Cooperation}. We will put our source code in the GitHub and leverage the whole community to improve the SDK for better performance and flexibility. More features, such as {{\color{Revision}python interface and PHP interface} can be expected with the support of the whole community.

\subsection{Implementation Details}
In this part, we will introduce the software architecture of the VIPLFaceNet. In Figure \ref{fig:sdkarc}, the architecture of VIPLFaceNet SDK is presented. Main
components in VIPLFaceNet SDK includes Blob, Command, Layer and Network Parser.

\textbf{Blob.} The blob is a container to hold the matrix in deep convolutional neural network. The Blob provides a mapping of the logic multi-dimensional matrix
to physical one-dimensional array.

\textbf{Command.} The command is an interface that provides basic network elements, e.g. convolution, rectifier linear unit(ReLU), max pooling or mean
pooling, and inner-product operation. As a SDK implementation for deployment, the implementation of the loss layers is unnecessary.

\textbf{Layer.} A VIPLFaceNet layer is the basic component of a deep neural network. In VIPLFaceNet, a layer is composed by one or more commands, e.g. a convolutional layer is composed by a convolution command and a ReLU command.

\textbf{Network Parser.} To facilitate the definition of network architecture, VIPLFaceNet SDK sets up a network parser. A network is defined by multiple layers and organized as a directed acyclic graph. The SDK code does not need to be modified regardless the network architecture changes.

\textbf{Matrix Multiplication Accelerating.} The single instruction multiple data (SIMD) instructions is used in VIPLFaceNet SDK to accelerate the matrix multiplication operation in both convolutional layer and fully-connected layer. Meanwhile, VIPLFaceNet also supports the third-party linear algebra library, e.g. Armadillo \cite{Armadillo}.

\section{Experiments and Analysis}
In this section, we compare the proposed VIPLFaceNet with the existing methods on the real-world face recognition benchmark LFW. Then, we discuss how VIPLFaceNet can be extended to other face recognition scenarios.

\subsection{Dataset and Evaluation Protocol}
The training set of our open-source VIPLFaceNet is the CASIA-Web dataset \cite{CASIA-Web}. All the training data are pre-processed as illustrated in section 3.3. {\color{Revision}Totally, we have 479,777 detected and normalized facial images of 10,575 identities in the training set}. The VIPLFaceNet is learned from scratch, {\color{Revision}and the weights of both the convolutional kernels and the fully-connected layers are initialized using the MSRA filler, i.e. $Var[w]=2/n$ , where $n$ is the number of input neurons \cite{PReLU}}. The mean of training images are subtracted firstly. During training, face patches equal to crop\_size $\times$ crop\_size pixel are randomly sampled from the input images and the images are also randomly flipped with 50\% probability. In all the experiments, the default crop\_size is set as 227.

The evaluation dataset is Labeled Faces in the Wild dataset (LFW) \cite{LFW}, which contains 13,233 images of 5,749 identities. For the standard 10-fold face verification experiment on LFW, we follow the unrestricted setting using external labeled data. {\color{Revision}Each fold of the test set consists of 300 inter-class and 300 intra-class face pairs}. We also conducted an additional experiment following the face identification protocol as in \cite{LFWOpenSet}.

\begin{table*}
\caption{The performance of our VIPLFaceNet and state-of-the-art methods on LFW View2 under the verification protocol.}
\vspace{-10pt}
\label{table:lfwverify}
\begin{center}
\begin{tabular}{lccc} \hline
Method                & Accuracy   & \# of Network & \# of Training Images \\ \hline
High-dim LBP\cite{HighdimLBP}          & 95.17\%    & --      &--\\
Fisher Vector Face\cite{FishervectorFace} & 93.03\%    & --      &--\\
DeepFace\cite{Deepface}                & 97.35\%    &3        &4M\\
DeepID\cite{DeepID}                    & 97.45\%    &25       &200K \\
DeepID2\cite{DeepID}                   & 99.15\%    &25      &200K \\
Gaussian Face\cite{GaussianFace}       & 98.52\%    & --      &--\\
DeepID2+\cite{DeepID2+}                & 99.47\%    &25       &290K\\
DeepID2+(Single)\cite{DeepID2+}        & 98.70\%    &1        &290K\\
WSTFusion\cite{WSTFusion}              & 98.37\%    &--       &10M \\
VGGFace\cite{VGGFace}                  & 98.95\%    &1        &2.6M\\
FaceNet\cite{FaceNet}                  & 99.63\%    &1        &200M  \\
AlexNet + FNL \cite{DCNN}                 & 97.70\%    &1        &500K\\
\textbf{VIPLFaceNetFull + FNL}        & \textbf{98.62\%}   &1        &500K\\
\textbf{VIPLFaceNet + FNL}    & \textbf{98.60\%}   &1        &500K\\ \hline
\end{tabular}
\end{center}
\vspace{-15pt}
\end{table*}

\subsection{Experimental Results on LFW}
In this part, we report the accuracy of VIPLFaceNet on LFW under both verification protocol and identification protocol. In all the experiments, we take the
2,048 outputs of the FC2 fully-connected layer as the representation of each face images and exploit the cosine function as the similarity metric between features.

\textbf{Comparisons with the state-of-the-art methods.}
In Table \ref{table:lfwverify}, we compare VIPLFaceNet with the state-of-the-art methods on LFW View 2 in terms of 10-fold mean accuracy. Using only single 10-layer deep convolutional neural network, our VIPLFaceNet achieves accuracy comparable to that of VGGFace \cite{VGGFace} with a 16-layer deep network and DeepID2+ with 25 deep networks, even though VIPLFaceNet is trained with less training data than DeepFace, VGGFace and FaceNet. VIPLFaceNet also reduces 40\% computation cost with slight performance degradation. Compared with AlexNet, VIPLFaceNet cuts down 40\% error rate on LFW (1.4\% vs. 2.3\%). The superiority of our method comes from the careful design of network architecture and simplification.

We further evaluate the VIPLFaceNet in the close-set and open-set face identification tasks, following the protocol in \cite{LFWOpenSet}. The close-set identification protocol reports the Rank-1 identification accuracy and the open-set identification reports the detection and identification rate (DIR) at False Alarm Rate (FAR) equal to 1\%. The comparisons with state-of-the-art methods are shown in Table \ref{table:lfwopen}. The proposed VIPLFaceNet also achieves state-of-the-art performance in the face identification tasks.

\begin{table}
\caption{The performance of our VIPLFaceNet and state-of-the-art methods on LFW under the identification protocol \cite{LFWOpenSet}.}
\vspace{-10pt}
\label{table:lfwopen}
\begin{center}
\begin{tabular}{llc} \hline
Method                       & Rank-1       & DIR $@$ FAR = 1\% \\ \hline
COTS-s1\cite{LFWOpenSet}     & 56.70\%      & 25.00\%      \\
COTS-s1+s4\cite{LFWOpenSet}  & 66.50\%      & 35.00\%      \\
DeepFace\cite{Deepface}      & 64.90\%      & 44.50\%      \\
WSTFusion\cite{WSTFusion}    & 82.50\%      & 61.90\%      \\
AlexNet+FNL\cite{DCNN}        & 89.26\%      & 58.72\%      \\
\textbf{VIPLFaceNetFull+FNL}      & \textbf{92.79\%}      &\textbf{68.13\%}      \\
\textbf{VIPLFaceNet+FNL}  & \textbf{91.95\%}      &\textbf{63.26\%}      \\ \hline
\end{tabular}
\end{center}
\vspace{-10pt}
\end{table}

\begin{table}
\caption{The performance of our VIPLFaceNet with different crop size on LFW View2 under the verification protocol.}
\vspace{-10pt}
\label{table:lfwcrop}
\begin{center}
\begin{tabular}{llc} \hline
Network Architecture  &Crop Size   & Accuracy  \\ \hline
VIPLFaceNet           &256         & 98.12\%          \\
VIPLFaceNet           &248         & 98.53\% \\
VIPLFaceNet           &\textbf{227}         & \textbf{98.60\%}          \\
VIPLFaceNet           &200         & 98.57\%    \\
VIPLFaceNet           &180         & 98.21\%    \\\hline
\end{tabular}
\end{center}
\vspace{-10pt}
\end{table}

\textbf{Comparisons of the crop size.} The crop size is related to both the performance and computational cost of deep networks. Due to the random data argumentation, the smaller the crop size, the bigger randomness in training. In Table \ref{table:lfwcrop}, we compare the performance of different crop size under the LFW verification protocol. It can be concluded that a moderate crop size 227 yields the best performance.

\textbf{Evaluation of the Fast Normalization Layer.} In Figure \ref{fig:nl}, we evaluate the effectiveness of the fast normalization layer (FNL) on LFW. As can be seen, adding FNL significantly improves the performance. Besides improving the generalization, FNL also significantly improves the convergence speed. By adding FNL, we can set the base learning rate as 0.04 and set the total echo as 15, while the baseline network needs 80 echoes for a good convergence. In Table \ref{table:lfwtraintime}, we compare the training time of VIPLFaceNet and VIPLFaceNet with FNL on the CASIA-Web training set. With the FNL, it only consumes 20\% training time and only slightly increases the online test cost.

\begin{figure}[t]
\centering
\includegraphics[width=8cm]{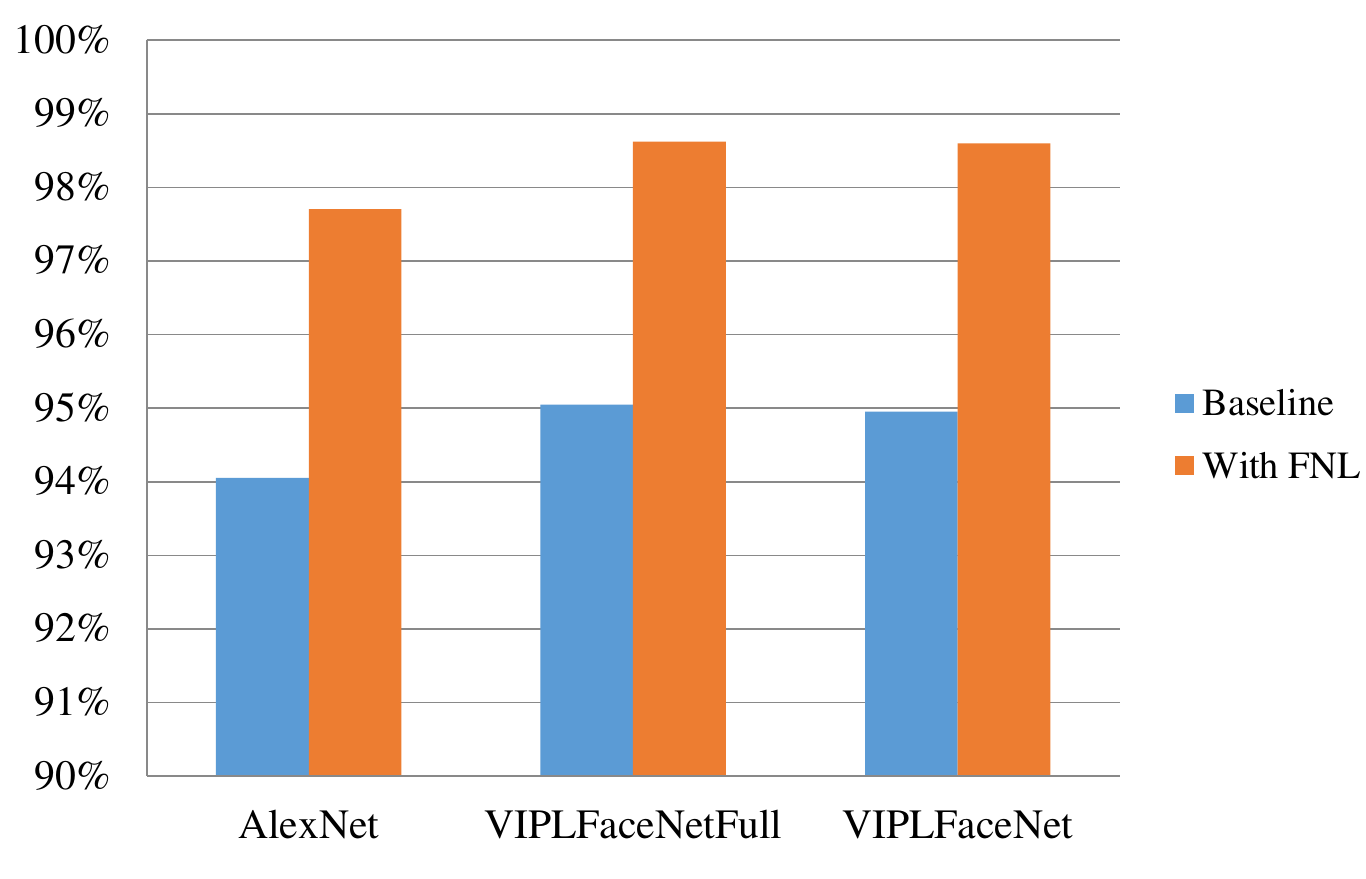}
\caption{The mean accuracy of our VIPLFaceNet on LFW View2 with or without FNL.}
\label{fig:nl}
\vspace{-10pt}
\end{figure}

\begin{table}
\caption{The time cost of our VIPLFaceNet with or without FNL. The training time of VIPLFaceNet with fast normalization layer is reduced by 80\%.}
\vspace{-10pt}
\label{table:lfwtraintime}
\begin{center}
\begin{tabular}{llc} \hline
Method                 & Training Time & Test speed on CPU  \\ \hline
AlexNet                & 67 hours      & 250ms / per image \\
VIPLFaceNetFull        & 60 hours      & 235ms / per image \\
VIPLFaceNet            & 40 hours      & 145ms / per image \\
VIPLFaceNetFull + FNL  & 12 hours     & 245ms / per image \\
VIPLFaceNet + FNL      & 8 hours      & 150ms / per image \\\hline
\end{tabular}
\end{center}
\vspace{-10pt}
\end{table}

\textbf{Extension of the VIPLFaceNet.} We can easily extend the VIPLFaceNet for more applications. For example, fine-tuning the VIPLFaceNet for other-domain applications such as age estimation \cite{AgeNet}.

\section{Conclusions}
In this paper, we propose and release an open-source deep face recognition SDK with carefully designed network architecture and simplification. By sticking a fast normalization layer to the ReLU layer, the training time is reduced by 80\% and the performance is significantly improved. On the real-world face recognition benchmark LFW, VIPLFaceNet achieves a mean accuracy of 98.60\%, which is comparable to the state-of-the-art. A fully C++ implementation of the {{\color{Revision}VIPLFaceNet SDK is released as an open source SDK under the BSD license}. VIPLFaceNet can serve as a good start point for both academic research and industrial applications under various real-world face recognition scenarios.

In the future, we would advocate the whole community to improve the SDK, e.g. supporting more language interface such as Python or PHP. Besides, we intend to build a VIPLFaceNet-based active face recognition development community and support more related application scenarios such as ID photo vs. real-word image verification, facial attribute analysis and age estimation.

\Acknowledgements{This work is partially supported by 973 Program under contract No. 2015CB351802, Natural Science Foundation of China under contracts Nos.  61402443, 61390511, 61379083, 61222211.}

\bibliography{PHL}
\bibliographystyle{fcs}


\Biography{xinliu}{Xin Liu recieved the B.S. degree at the Chongqing University, Chongqing, China, in June 2011. Currently, he is a Ph.D candidate at the Institute of Computing Technology, Chinese Academy of Sciences. His research interests include face recognition, image retrieval and deep learning.}

\Biography{mnkan}{Meina Kan is an Associate Professor with the Institute of Computing Technology (ICT), Chinese Academy of Sciences (CAS). She received the Ph.D. degree from the University of Chinese Academy of Sciences (CAS), Beijing, China. Her research mainly focuses on Computer Vision especially face recognition, transfer learning, and deep learning.}

\Biography{wlw}{Wanglong Wu recieved the B.S. degree at the Beijing Jiaotong University, Beijing, China, in June 2014. Currently, he is a Ph.D candidate at the Institute of Computing Technology, Chinese Academy of Sciences. His research interests include face recognition and deep learning.}

\Biography{sgshan}{Shiguang Shan received M.S. degree in computer science from the Harbin Institute of Technology, Harbin, China, in 1999, and Ph.D. degree in computer science from the Institute of Computing Technology (ICT), Chinese Academy of Sciences (CAS), Beijing, China, in 2004. He joined ICT, CAS in 2002 and has been a Professor since 2010. He is now the Deputy Director of the Key Lab of Intelligent Information Processing of CAS. His research interests cover computer vision, pattern recognition, and machine learning. He especially focuses on face recognition related research topics. He has published more than 200 papers in refereed journals and proceedings.}

\Biography{xlchen}{Xilin Chen received the B.S., M.S., and Ph.D. degrees in computer science from the Harbin Institute of Technology, Harbin, China, in 1988, 1991, and 1994, respectively. He is a Professor with the Institute of Computing Technology, Chinese Academy of Sciences (CAS). He has authored one book and over 200 papers in refereed journals and proceedings in the areas of computer vision, pattern recognition, image processing, and multimodal interfaces.}

\begin{minipage}{\columnwidth}

\end{minipage}

\end{document}